\documentclass[10pt,twocolumn,letterpaper]{article}

\usepackage{wacv}
\usepackage{times}
\usepackage{epsfig}
\usepackage{graphicx}
\usepackage{amsmath}
\usepackage{amssymb}
\usepackage{booktabs}
\usepackage{color}
\usepackage{multirow}
\usepackage{graphicx}
\usepackage{multirow}
\usepackage{graphicx,multirow}
\usepackage{wrapfig}
\usepackage{xcolor}
\usepackage{bm}
\usepackage{booktabs}
\usepackage{adjustbox}
\usepackage{pifont}
\usepackage{subcaption}
\newcommand{\xmark}{\ding{55}}
\usepackage[accsupp]{axessibility}  

\def\wacvPaperID{825} 

\wacvalgorithmstrack   

\wacvfinalcopy 

\ifwacvfinal
\usepackage[breaklinks=true,bookmarks=false]{hyperref}
\else
\usepackage[pagebackref=true,breaklinks=true,colorlinks,bookmarks=false]{hyperref}
\fi

\begin{document}

\title{DDAM-PS: Diligent Domain Adaptive Mixer for Person Search}

\author{Mohammed Khaleed Almansoori*$^1$ \quad Mustansar Fiaz*$^{1,2}$ \quad Hisham Cholakkal$^{1}$ \\
$^1$Mohamed bin Zayed University of Artificial Intelligence, Abu Dhabi, UAE \hspace{1.5mm} $^2$IBM   \\
{\tt\small (mohammed.almansoori, hisham.cholakkal)@mbzuai.ac.ae  \hspace{1.5mm}  mustansar.fiaz@ibm.com}
 }


\maketitle

\begin{abstract}
Person search (PS) is a challenging computer vision problem where the objective is to achieve joint optimization for pedestrian detection and re-identification (ReID). Although previous advancements have shown promising performance in the field under fully and weakly supervised learning fashion, there exists a major gap in investigating the domain adaptation ability of PS models. In this paper, we propose a diligent  domain adaptive mixer (DDAM) for person search (DDAP-PS) framework that aims to bridge a gap to improve knowledge transfer from the labeled source domain to the unlabeled target domain. Specifically, we introduce a novel DDAM module that generates moderate mixed-domain representations by combining source and target domain representations. The proposed DDAM module encourages domain mixing to minimize the distance between the two extreme domains, thereby enhancing the ReID task. To achieve this, we introduce two bridge losses and a disparity loss. The objective of the two bridge losses is to guide the moderate mixed-domain representations to maintain an appropriate distance from both the source and target domain representations. The disparity loss aims to prevent the moderate mixed-domain representations from being biased towards either the source or target domains, thereby avoiding overfitting. Furthermore, we address the conflict between the two subtasks, localization and ReID, during domain adaptation. To handle this cross-task conflict, we forcefully decouple the norm-aware embedding, which aids in better learning of the moderate mixed-domain representation. We conduct experiments to validate the effectiveness of our proposed method. Our approach demonstrates favorable performance on the challenging PRW and CUHK-SYSU datasets.  Our source code is publicly available at \url{https://github.com/mustansarfiaz/DDAM-PS}. 
\end{abstract}
\def\thefootnote{*}\footnotetext{These authors contributed equally to this work.}
\def\thefootnote{$\ddagger$}\footnotetext{Mustansar’s work on this paper was done when he was at MBZUAI.}

\begin{figure}[t]
   \centering
         \includegraphics[width=0.8\linewidth]{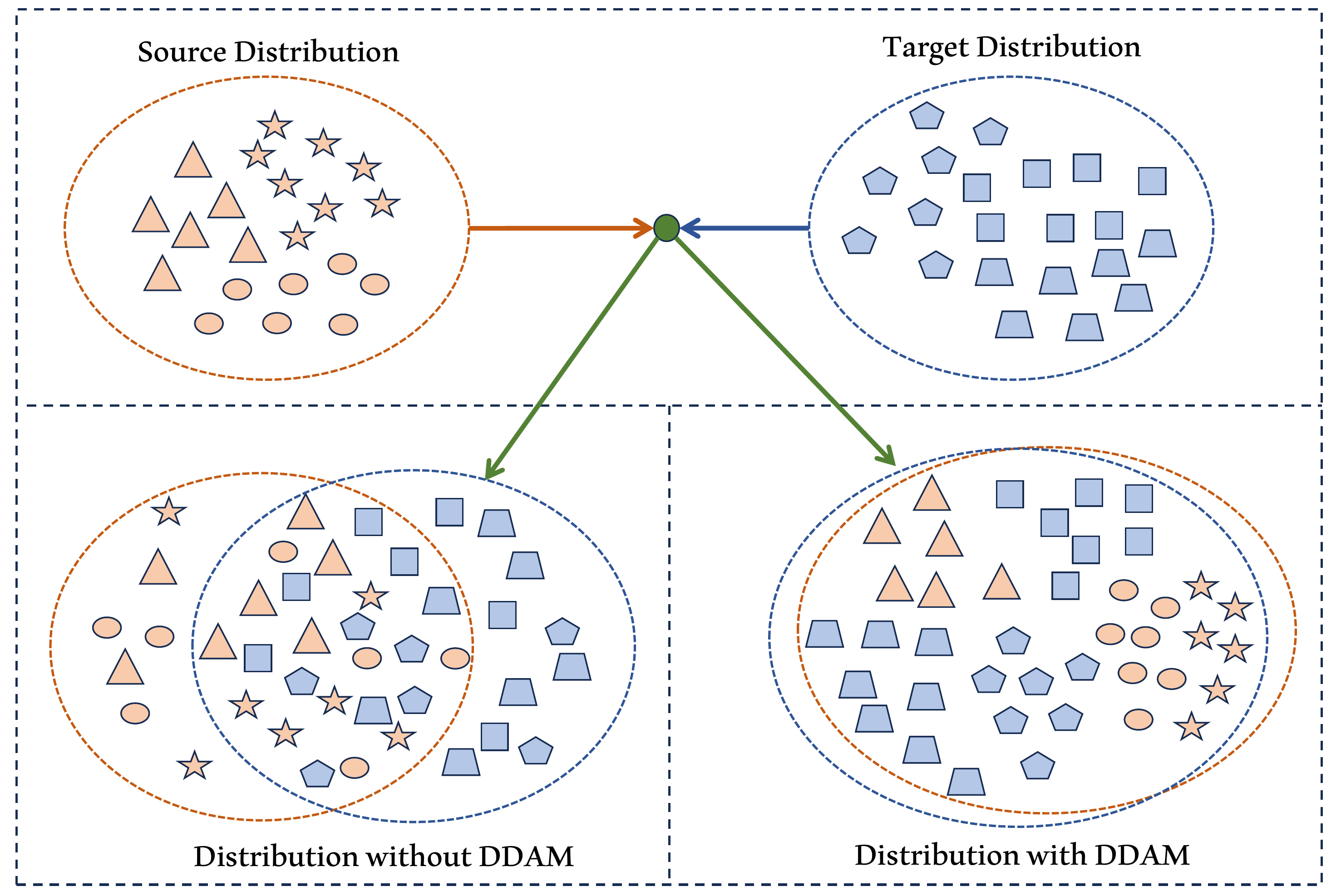}
    \caption{
    Demonstration of the impact of domain adaption with and without our proposed diligent  domain adaptive mixer (DDAM) module for the person search problem. Suppose, the source and target feature points are localized in hyperspace. In order to better transfer the source knowledge to the target domain, our proposed DDAM finds moderate mixed-domain distribution to bridge the gap between the source and target distributions. Here various shapes and colors donate the different distributions and different person identities correspondingly.    
    }
    \label{fig:intro_fig}
\end{figure}

\section{Introduction}
Person search aims to optimize two conflicting subtasks: detection and re-identification (ReID) \cite{oimnet,psarm,nae}. Detection focuses on localizing pedestrians in a given scene, while ReID is responsible for uniquely identifying individuals. This research problem becomes extremely complex due to the utilization of real-world data sources (such as CCTV), which often contain uncropped images with varying specifications, resolutions, lighting conditions, and other variations. While person search has been extensively explored under the fully supervised learning \cite{oimnet,psarm,pstr,alignps}  and weakly supervised learning \cite{WSPS_1, WSPS_2} paradigms, adapting it for unsupervised domain adaptation (UDA) generalization remains challenging, as there is a significant disparity between the distributions of the source and target domains.

Unsupervised domain adaptation (UDA) has demonstrated promising results in various domains, including aerial tracking  \cite{ye2022unsupervised}, nighttime semantic segmentation  \cite{gao2022cross}, visual recognition \cite{yang2023tvt, huang2022category,zhang2022spectral}, and person ReID \cite{zheng2022unsupervised, mohanty2022ssmtReid,cheng2022h}. Unlike fully supervised learning and weakly supervised learning, UDA focuses on bridging the gap between the ideal training set and real-world scenarios by leveraging labeled source data and transferring learned knowledge to unlabeled target domains. Li et al. \cite{daps} are the first to apply UDA to person search and proposed DAPS, a method that employs implicit alignment modules and pseudo-labeling to reduce the discrepancy between source and target domains.
However, DAPS suffers from a lack of an explicit bridge to determine which critical information, such as similarity or dissimilarity, should be utilized to mitigate the domain discrepancy. Moreover, the implicit alignment modules employed in challenging real-world scenarios, where the person search (PS) model encounters scene challenges like occlusion and pose variations, as well as environmental challenges such as diverse indoor and outdoor scene distributions, may deteriorate the region of interest.
Existing PS \cite{psarm, nae, seqnet} methods based on Faster-RCNN \cite{fasterrcnn}  strive to jointly optimize the conflicting subtasks of detection and ReID. In an effort to address this issue, Chen et al. \cite{nae} introduced norm-aware embedding (NAE) to disentangle the two tasks. However, it still utilizes shared weights for both detection and ReID. Therefore, directly utilizing shared NAE representations for domain adaptation may further increase the complexity of person search.

To address the challenges mentioned above, we propose a diligent domain adaptive bridging mechanism to learn domain-invariant feature representations by introducing a bridge that reduces or minimizes the discrepancy between the two domains. Inspired by \cite{IDM}, we aim to enhance knowledge transfer between the source and target domains by learning mixed-domain representations from both domains.
As discussed earlier, a significant domain shift exists between the distributions of the two domains. In Fig. \ref{fig:intro_fig}, we illustrate the region of interest (RoI) proposals from the source and target distributions in hyperspace. Our bridging mechanism introduces hidden representations, referred to as moderate mixed-domain representations, with the objective of smoothly transferring RoI knowledge from the source domain to the target domain. To achieve this, we enforce two bridge losses on the moderate domain representations, minimizing the distance between the source and target domain representations.
Additionally, we employ a disparity loss that regularizes the diversity between the two domains by maximizing the standard deviation. This regularization helps to avoid overfitting to either of the domains and facilitates gradual domain adaptation. Depending on the ambient nature of the mixed-domain representations, the source RoI labels can dominate or the inherent distribution of the target domain can be more exposed. The bridge losses and disparity loss work together to learn mixed-domain representations, allowing the model to effectively transfer source RoI knowledge and enhance discriminability in the target domain for the ReID task.
Furthermore, we propose to decouple the norm-aware embeddings to mitigate the conflict between detection and ReID, which in turn simplifies the process of domain adaptation. Through experiments, we demonstrate that our approach surpasses the state-of-the-art method DAPS on the PRW and CUHK-SYSU datasets.

\noindent\textbf{Contribution:} 
Our contributions can be summarized as follows:
(1) We propose an explicit diligent domain adaptive mixing mechanism to reduce the gap between the source and target domains in the person search domain adaptation problem. Specifically, we learn mixed domain representations that bridge the discrepancy between the two domains and facilitate the swift transfer of source information to the target domain, thereby promoting UDA person search tasks.
(2) To enhance domain adaptation ability and generate elegant mixed domain representations, we introduce two bridge losses and a disparity loss.
(3) To alleviate the conflict between detection and ReID and further improve domain adaptation, we propose the decoupling of the NAE representation.
(4) Experimental results demonstrate the promising performance of our method on two datasets, outperforming state-of-the-art methods. These results highlight the merits of our approach.

\section{Related Work}
\subsection{Person Search} Person search aims to unify the sub-tasks of localization of pedestrians \cite{cao2021handcrafted, Pang_MGAN_ICCV_2019, fasterrcnn} and re-identification of the person of interest \cite{Ye_ReIDSurvey_TPAMI_2020, gu2022clothes, li2021diverse} in an end-to-end model. The PS problem becomes a popular research topic, and methods start to focus on the challenges of the two contradictory objectives. The challenge comes when pedestrian detection aims to extract common features to improve localization, while ReID pushes to extract unique features of the same individual.
PS problem can be classified as two-stage \cite{Lan_CLSA_ECCV_2018,Dong_IGPN_CVPR_2020,girshick2015deformable,ps_mask} and one-stage \cite{nae,psarm,fiaz2023sat,coat} methods.  In two-stage methods, first detection is performed to locate the pedestrians employing off-the-shelf detectors, and later re-identification task is performed over the cropped pedestrians for identity discrimination. Although two-stage methods provide promising performance, they face immense computational costs.

On the contrary, one-stage methods perform both sub-tasks simultaneously in an end-to-end manner. 
These one-stage methods exploit the two-stage detector i.e., Faster RCNN \cite{fasterrcnn}, and combine additional ReID loss for pedestrian identity discrimination. For example,  OIM \cite{oimnet,PRW} utilized Faster RCNN to implement an end-to-end Person search model.   NAE \cite{nae} disentangle the detection and the ReID into a norm and angle Euclidian representation, allowing to minimize the cross-task conflict. Furthermore, inherited disadvantages of Faster-RCNN affect the gains for PS, thus sequential models \cite{seqnet,coat,psarm,fiaz2023sat,oimnet++} mitigate the low-quality proposal of the RPN. Seqnet \cite{seqnet} sequential structure allowed the model to focus on reducing the cross-task conflict by getting a better proposal and for the final stage to focus more on the ReID. The COAT \cite{coat} utilized transformer encoders to shuffle patches of individuals with each other in other to generalize better for unseen images. Studies such as PS-ARM \cite{psarm} introduce the attention-aware relation mixer   to exploit the relations between different local regions within RoI of
a person.

The works \cite{pstr,alignps} motivate to further disentangle the two-sub tasks, by moving away from Faster R-CNN structure due to limitation and computational resources. To address these issues, AlignPS \cite{alignps} uses anchor free approach to eliminate the need for using low-quality proposals. In addition, utilizing an aligned feature aggregation module  mitigates the issues of scale, region, and task alignment. Cao et al. \cite{pstr} introduces Deformable Detr \cite{zhu2021deformable} for PS that simultaneously predicts the detection and ReID. However, these fully supervised methods (FSL) methods suffer from the issue of domain gap which, degrades the performance of the model. To minimize the domain shift issue, recent studies in  weakly supervised person search (WSPS) \cite{WSPS_1,WSPS_2} have access to bounding boxes with ID annotations. Although these issue helps to reduce the domain gap, they still require label data. Another recent study is DAPS \cite{daps} which introduces the concept of UDA in PS. DAPS focuses on the domain alignment between the source and the target domain, and also the pseudo-labeling framework for the target domain.     
In contrast, we propose a novel bridging mechanism  that enhances the discriminative learning for ReID by bridging the gap between the source and target domains as well as minimizing the cross-task conflict with localization to ease the domain generalization task.

\subsection{Domain Adaptation for Person ReID}
Unsupervised domain adaptation (UDA) for person ReID approaches are exposed to labeled source domain and translate the learned knowledge to the underlying target domain in an unsupervised manner. The UDA person ReID approaches are classified into three categories based on their training strategies including GAN transferring \cite{deng2018image, wei2018person}, joint training \cite{zhong2020learning, ge2020self}, and fine-tuning \cite{dai2021dual, fu2019self}. GAN transferring approaches utilize GAN models to disentangle the style discrepancy and  transfer the learned information from the source to the target domain.  For joint training, the approaches employ a memory bank that combines the source and target data and jointly trains without building a bridge between the two domains to improve the target domain features.  However, for fine-tuning methods, they train the model for source data and fine-tune over target data using pseudo labels. The key component is to mitigate the effect of noisy pseudo labels. 
Nevertheless, UDA person ReID is based on cropped pedestrians and cannot be directly extended for the person search problem. The DAPS \cite{daps} proposed a  clustering mechanism to provide high-quality pseudo labels to expedite the target domain training. However, DAPS implicitly utilizes the source and target data while ignoring the explicit bridge mechanism to alleviate the gap between the two domains. Therefore, inspired by \cite{IDM}, we introduced an explicit mechanism to learn what similar/dissimilar information can be employed to improve the target domain features for ReID.

\begin{figure*}[t]
   \centering
    \includegraphics[width=0.9\linewidth]{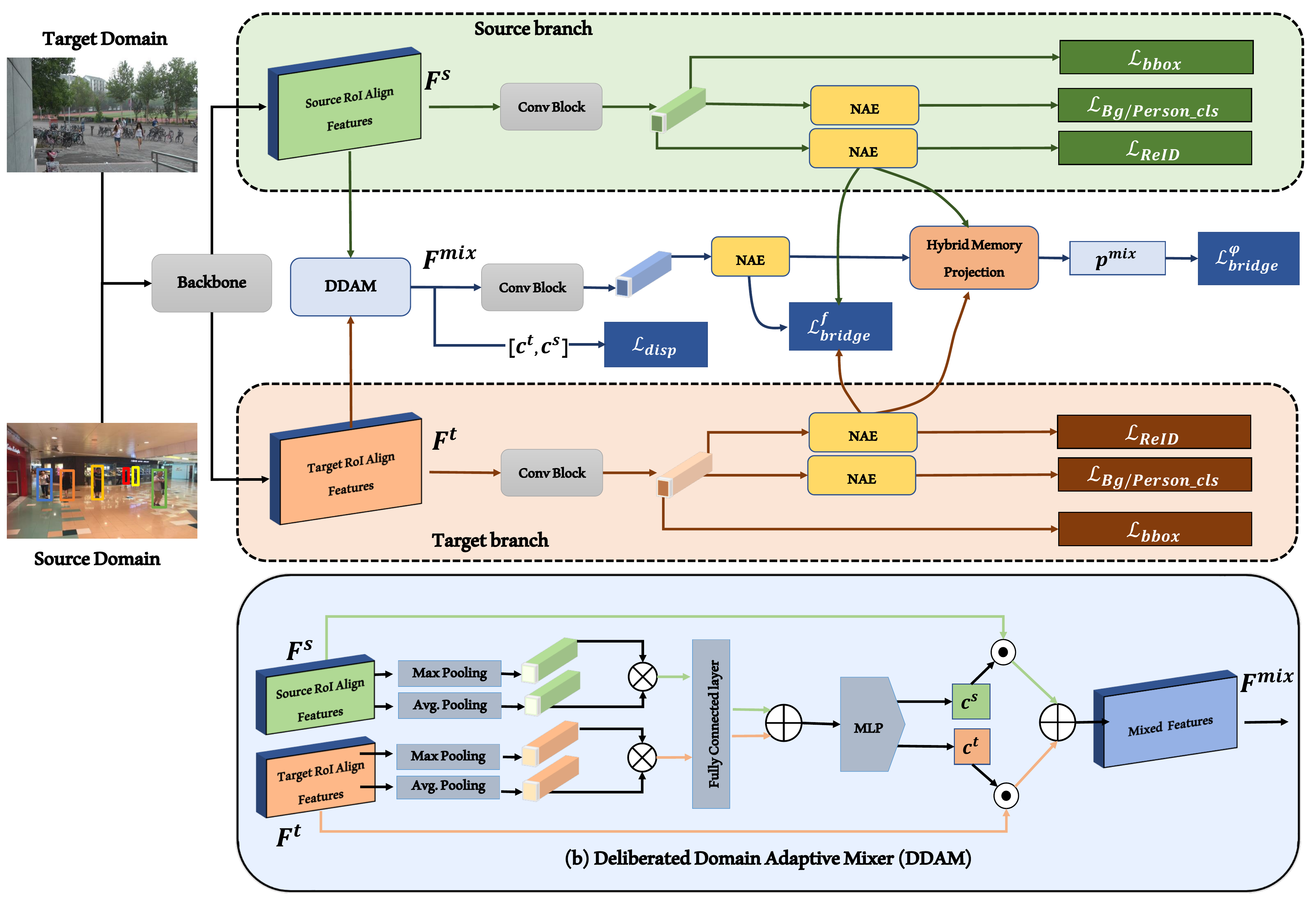}
    \caption{The illustration of our proposed diligent domain adaptive mixer person search (DDAM-PS) framework. The source and target stem features are computed using a backbone and input to the RPN \cite{fasterrcnn} to compute RoI align features. These source and target RoI align features ($F^s, F^t$) are fed to the diligent  domain adaptive mixer (DDAM) module to generate the mixed domain representations ($F^{mix}$), to reduce the domain gap for unsupervised domain adaptation (UDA), as shown in (b).  To generate moderate mixed domain representations, we employ two bridges losses ($\mathcal{L}^f_{bridge}$ and $\mathcal{L}^{\varphi}_{bridge}$) and a disparity loss ($\mathcal{L}_{disp}$). The $\mathcal{L}^f_{bridge}$  loss is applied using the NAE embedding of the target-domain, source-domain, or mixed-domain representations to evaluate the distance across the domains. While $\mathcal{L}^{\varphi}_{bridge}$ is enforced using a hybrid memory projection module   to measure the discrepancy between the mixed domain memory projections and the two domains. 
    The disparity loss is enforced to regulate the mixed domain features, to avoid overfitting using constraint weights ($c^s, c^t$), obtained from the DDAM module.
    In addition, we propose to decouple the NAE module and apply separate NAE for both conflicting subtasks i.e., detection and ReID. This decoupling facilitates to adopt it for the UDA ReID task. }    
   \label{fig_proposed_framework}
\end{figure*}

\section{Method}

The overall framework of our proposed diligent domain adaptive mixer for person search, DDAM-PS, is illustrated in Fig. \ref{fig_proposed_framework}. It jointly takes input from both the source and target domains. The base network of our framework is DAPS \cite{daps}, which incorporates an implicit domain alignment module (DAM) to reduce the gap between the two domains. The source and target domain images are fed into a ResNet50 \cite{he2016deep} backbone network to extract feature embeddings. These embeddings are then input to the region proposal network (RPN) \cite{fasterrcnn} to generate ROI-Aligned proposal candidates. To enhance the ReID task within the baseline, we introduce a diligent domain adaptive mixing (DDAM) mechanism. This mechanism aims to smooth out the extreme differences between the source and target domains, allowing for better domain adaptation.
To achieve this, we fuse the source and target domain proposals to generate new mixed domain proposal representations.
{For the detection task, we employ a combination of box regression head ($\mathcal{L}_{bbox}$) and person vs background classification head ($\mathcal{L}_{Bg/Person\_cls}$) to compute detection losses.}
For the ReID task, we impose OIM \cite{oimnet} ReID loss, denoted as $\mathcal{L}_{ReID}$, on the source and target domain features. Pseudo-labels for the target domain are generated using a clustering strategy.
In order to generate moderate mixed-domain adaptive representations, we introduce two bridge losses, $\mathcal{L}^{f}_{bridge}$ and $\mathcal{L}^{\varphi}_{bridge}$.
The $\mathcal{L}^f_{bridge}$ loss is applied by utilizing the NAE embedding of the target-domain, source-domain, or mixed-domain representations to evaluate the distance across the domains. On the other hand, $\mathcal{L}^{\varphi}_{bridge}$ is enforced using a hybrid memory projection module to measure the discrepancy between the mixed domain memory projections and the two domains.
Additionally, we employ a disparity loss to regulate the domain mixing mechanism and prevent overfitting to either of the two extreme domains.
As mentioned earlier, person search models face challenges in jointly optimizing the two subtasks of object detection and ReID. When adapting these models for UDA, the complexity further increases. To address this issue, we propose to decouple the norm-aware embeddings (NAE). This decoupling not only alleviates the conflict between the two subtasks but also improves the PS domain adaptation framework.


\subsection{Diligent  Domain Adaptive Mixer (DDAM)}
\label{IDM_section}

Inspired by \cite{IDM}, we propose an explicit mixed domain representation learning approach to enhance knowledge transfer between source and target domains for UDA PS.
The DDAM module takes $n$ pairs of RoI pooled features from both the source ($F^s$) and target ($F^t$) domains and generates domain constraint weights, denoted as $c^s$ and $c^t$, respectively. The source ($F^s$) and target ($F^t$) RoI features are realized with the average and maximum pooling operations. These pooled features are then concatenated for each domain and passed through a shared fully connected (FC) layer. The features from the FC layer are merged via element-wise summation operation and input to a multi-layer perceptron (MLP) followed by a Softmax activation function, yielding the domain constraint weights. The overall procedure to obtain the domain constraint weights is illustrated in Figure \ref{fig_proposed_framework}-(b). The two domain constraint weights $c^s$ and $c^t$ are represented as [$c^s$, $c^t$] = $c$, where $c \in \mathbb{R}^2$.
Finally, the RoI mixed domain representations are achieved by mixing the source RoI features and target RoI features using the two domain constraint weights as follows:  
\vspace{-0.2cm}
\begin{equation}\label{eq_d_weights}
    F^{mix}=c^s\cdot F^s+c^t\cdot F^t.
\end{equation}

\begin{table}[t!]
\begin{center}
  \caption{The quantitative comparison of our's unsupervised domain adaptive (UDA) method with fully supervised state-of-the-art methods on both the CUHK-SYSU and PRW datasets. The performance is evaluated using mAP and top-1 accuracy.  Our method scores are in bold.}
  \label{tab:tab_sota_comparison}
\scalebox{0.85}{\begin{tabular}{|cl|cc|cc|c|}
\hline
\multicolumn{2}{|c|}{}  & \multicolumn{2}{|c|}{CUHK-SYSU}   & \multicolumn{2}{|c|}{PRW}    \\ \cline{3-6}
\multicolumn{2}{|c|}{\multirow{-2}{*}{Method}} & \multicolumn{1}{|c|}{mAP} & \multicolumn{1}{|c|}{top-1} & \multicolumn{1}{|c|}{mAP} & \multicolumn{1}{|c|}{top-1} \\ \cline{3-6} \hline \hline
\multicolumn{1}{|c|}{  \multirow{6}{*}{\rotatebox[origin=c]{90}{Two-step}}}  & CLSA \cite{Lan_CLSA_ECCV_2018} & \multicolumn{1}{c|}{87.2} & 88.5& \multicolumn{1}{c|}{38.7} & 65.0  \\
\multicolumn{1}{|c|}{} & IGPN \cite{Dong_IGPN_CVPR_2020}   & \multicolumn{1}{c|}{90.3}  & 91.4 & \multicolumn{1}{c|}{42.9} & 70.2   \\ 
\multicolumn{1}{|c|}{} & DPM \cite{girshick2015deformable}  & \multicolumn{1}{c|}{-}  &   -    & \multicolumn{1}{c|}{20.5}     &  48.3     \\ 
\multicolumn{1}{|c|}{} & RDLR \cite{ps_local_refi}   & \multicolumn{1}{c|}{93.0}     &  94.2 & \multicolumn{1}{c|}{42.9}     &   70.2      \\ 
\multicolumn{1}{|c|}{} & MGTS  \cite{ps_mask} & \multicolumn{1}{c|}{83.0}   &83.7 & \multicolumn{1}{c|}{32.6}  &   72.1     \\
\multicolumn{1}{|c|}{} & TCTS \cite{Cheng_TCTS_CVPR_2020}  & \multicolumn{1}{c|}{93.9}  & 95.1 & \multicolumn{1}{c|}{46.8}     &  87.5   \\
\hline
\multicolumn{1}{|c|}{  \multirow{20}{*}{\rotatebox[origin=c]{90}{End-to-end}}} & OIM  \cite{oimnet} & \multicolumn{1}{c|}{75.5} & 78.7  & \multicolumn{1}{c|}{21.3} & 49.9 \\
\multicolumn{1}{|c|}{} & RCAA \cite{Chang_RCAA_ECCV_2018}   & \multicolumn{1}{c|}{79.3} & 81.3   & \multicolumn{1}{c|}{-}  & -       \\ 
\multicolumn{1}{|c|}{} & NPSM \cite{Liu_NPSM_ICCV_2017}   & \multicolumn{1}{c|}{77.9}     & 81.2    & \multicolumn{1}{c|}{24.2}     &  53.1        \\
\multicolumn{1}{|c|}{} & IAN \cite{Xiao_IAN_PR_2019}   & \multicolumn{1}{c|}{76.3}     & 80.1    & \multicolumn{1}{c|}{23.0}     & 61.9     \\
\multicolumn{1}{|c|}{} & QEEPS \cite{Munjal_QEEPS_CVPR_2019}   & \multicolumn{1}{c|}{88.9}     & 89.1    & \multicolumn{1}{c|}{37.1}     & 76.7      \\
\multicolumn{1}{|c|}{} & CTXGraph \cite{Yan_CTXG_CVPR_2019}   & \multicolumn{1}{c|}{84.1}     & 86.5    & \multicolumn{1}{c|}{33.4}     &73.6      \\
\multicolumn{1}{|c|}{} & HOIM \cite{chen2020hierarchical}   & \multicolumn{1}{c|}{89.7}     & 90.8   & \multicolumn{1}{c|}{39.8}     &  80.4     \\
\multicolumn{1}{|c|}{} & BINet \cite{ps_bi-dir}   & \multicolumn{1}{c|}{90.0}     & 90.7   & \multicolumn{1}{c|}{45.3}     & 81.7   \\
\multicolumn{1}{|c|}{} & APNet \cite{Zhong_APNet_CVPR_2020}   & \multicolumn{1}{c|}{88.9}     & 89.3   & \multicolumn{1}{c|}{41.2}     & 81.4     \\
\multicolumn{1}{|c|}{} & AlignPS \cite{Yan_AlignPS_CVPR_2021}   & \multicolumn{1}{c|}{93.1}     & 93.4   & \multicolumn{1}{c|}{45.9}     & 81.9      \\
\multicolumn{1}{|c|}{} & AlignPS + \cite{alignps} & \multicolumn{1}{c|}{94.0}     & 94.5   & \multicolumn{1}{c|}{46.1}     & 85.8    \\
\multicolumn{1}{|c|}{} & NAE \cite{alignps}   & \multicolumn{1}{c|}{91.5}     & 92.4   & \multicolumn{1}{c|}{43.3}     & 80.9    \\
\multicolumn{1}{|c|}{} & SeqNet  \cite{seqnet}   & \multicolumn{1}{c|}{93.8} & 94.6   & \multicolumn{1}{c|}{46.7}     & 83.4     \\
\multicolumn{1}{|c|}{} &  PSTR \cite{pstr}   & \multicolumn{1}{c|}{93.5} & 95.0 & \multicolumn{1}{c|}{49.5} & 87.8 \\
\multicolumn{1}{|c|}{} &  OIMNet++ \cite{oimnet++}   & \multicolumn{1}{c|}{93.1} & 93.9 & \multicolumn{1}{c|}{46.8} & 83.9 \\

\cline{1-6}
\multicolumn{1}{|c|}{  \multirow{1}{*}{UDA}} &
\textbf{Ours}   & \multicolumn{1}{c|}{\textbf{79.5}} & \textbf{81.3} & \multicolumn{1}{c|}{\textbf{36.7}} & \textbf{81.2} \\

\hline
\end{tabular}}
\end{center}
\end{table}

\begin{table}[t!]
\centering
\caption{Comparison of our method with weakly supervised methods and domain adaptive state-of-the-art PS methods over PRW and CUHK-SYSU datasets. The * indicates the training of R-SiamNet using both CUHK-SYSU and PRW.  The best results are in bold.}
\label{tbl:comparison_with_semI_DA}
\setlength{\tabcolsep}{8pt}
\adjustbox{width=\columnwidth}{
\begin{tabular}{|l|l|c|c|c|c|}
\cline{1-6}
\multicolumn{2}{|c|}{\multirow{2}{*}{Methods}} & \multicolumn{2}{c|}{PRW}                              & \multicolumn{2}{c|}{CUHK-SYSU}                        \\ \cline{3-6} 
\multicolumn{2}|{l|}{}                        & \multicolumn{1}{|l|}{mAP} & \multicolumn{1}{l|}{top-1} & \multicolumn{1}{|l|}{mAP} & \multicolumn{1}{l|}{top-1} \\ \cline{1-6}  \hline
\multirow{3}{*}{Weakly-Supervised}     & CGPS \cite{WSPS_1} & 16.2 & 68.0  & 80.0 & 82.3 \\
                                     & R-SiamNet \cite{WSPS_2} &  21.4 & 75.2 & 86.0 & 87.1 \\
                                     & R-SiamNet$*$ \cite{WSPS_2} & 23.5 & 76.0 & 86.2 & 87.6 \\  \hline
\multirow{2}{*}{UDA}   & DAPS \cite{daps} &  34.7 & 80.6 & 77.6 & 79.6 \\   \cline{2-6}     
                                     & \textbf{Ours} &  \textbf{36.7} & \textbf{81.2} & \textbf{79.5} & \textbf{81.3}   \\ \hline 
\end{tabular}}
\end{table}

\begin{figure}[t!]
  \centering    \includegraphics[width=0.48\textwidth]{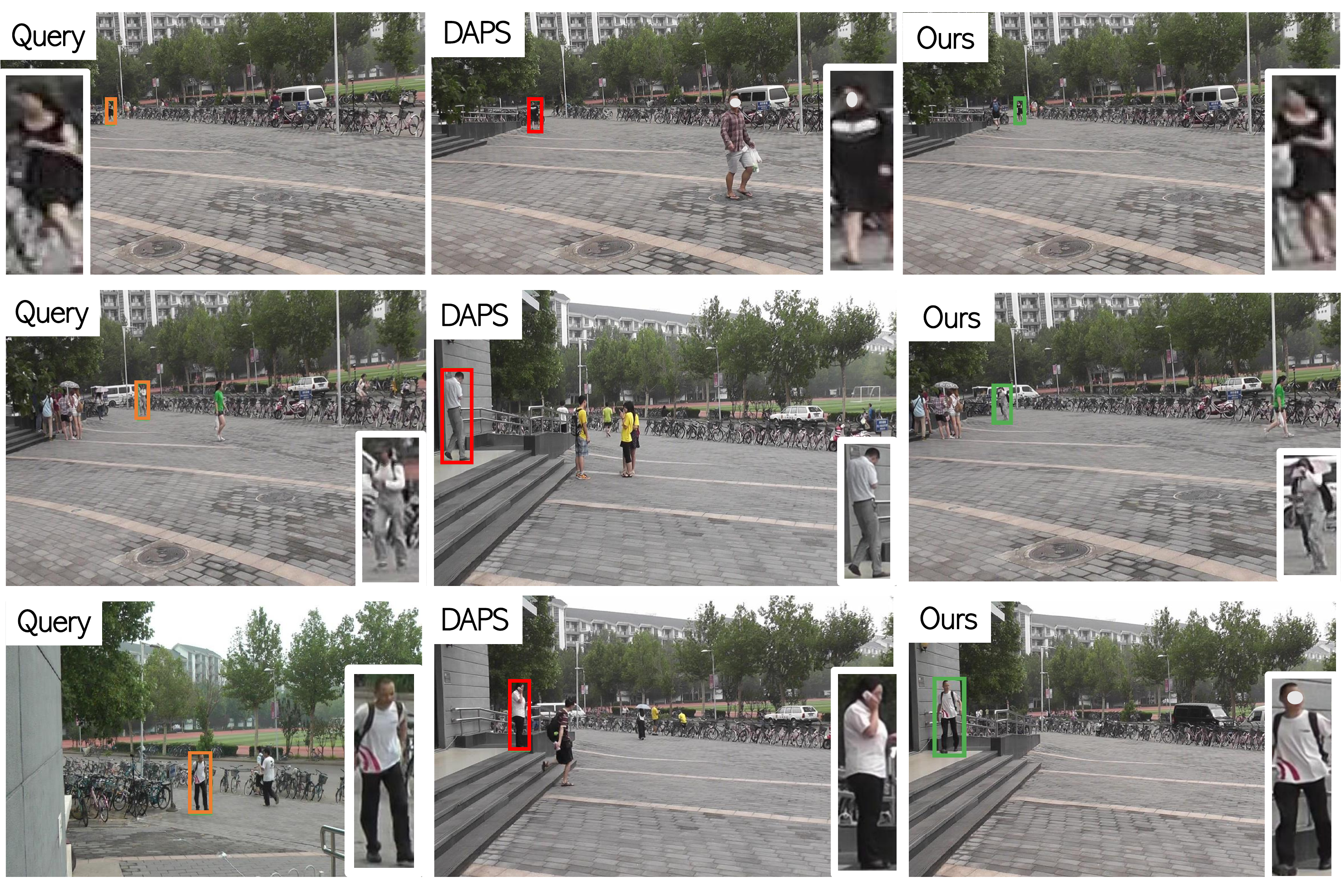}
  \caption{Qualitative comparison between the DAPS \cite{daps} and our's approach in three different challenging scenes. Our method predicts correct top-1 matching results. The orange, red, and green colors show the query, failure, and correct, respectively.}
  \label{fig:daps_prw_qualitative}
\end{figure}

\subsection{Moderate Domain Mixing}


The effectiveness of domain mixing can be hindered by two factors:
(1) The RoI pooled feature samples often exhibit diverse backgrounds, and individuals within both the intra-domain and inter-domain distributions may experience appearance variations. This includes challenges related to environmental factors such as indoor and outdoor scenes.
(2) From Equation \ref{eq_d_weights}, we can generate an infinite number of mixed domain representations by exposing  source and target domain's RoI features to the DDAM module. However, only a limited portion of these mixed domain representations is capable of effectively bridging the gap between the two extreme domains. These factors can potentially degrade the quality of the mixed domain representations.

In order to better learn the mixed domain  distribution ($P_{mix}$), the source distribution ($P_s$) and target distribution ($P_t$)  should be located on the shortest path \cite{gopalan2013unsupervised} (see Fig. \ref{fig:intro_fig}).
Although the baseline learns the domain-invariant representations using the DAM module, 
this approach does not take into account the extreme classes for each domain which is  more likely to affect the class distributions in each domain.
Therefore, considering the shortest distance definition,  the mixed domain representations should follow the two desired characteristics which are ensured by enforcing the dedicated losses. 
To bridge the extreme domains in the hyperspace,  the distance $d(.)$ should be proportional where $c^s+c^t=1$ {(using softmax function)} and $c^s,c^t \in [0,1]$. Thus, the moderate mixed domain representation can be obtained  utilizing domain constraint weights by identifying the closest points to both  $P^s$ and $P^t$ as well as localized along the shortest path.  The problem can be framed as loss minimization as follows:
\vspace{-0.2cm}
\begin{equation}
    \mathcal{L}_{bridge}=c^{s} \cdot d(P_s,P^{(c)}_{mix}) + c^{t} \cdot d(P_t,P^{(c)}_{mix}).
\label{a_eq}
\end{equation}

The enforced loss (Eq. \ref{a_eq}) controls the gap between two domains by minimizing the shift between the two domains. The $\mathcal{L}_{bridge}$ loss will punish more $ d(P_t,P^{(c)}_{mix})$ if $c^t > c^s$, else it will  push more $ d(P_s,P^{(c)}_{mix})$. The domain constraint weights ($c^t, c^s$)  in DDAM  ensure a steady domain adaptive procedure to balance the minimization of the domain shifts from the source to the target domains.

We impose bridge losses on the mixed domain feature representations and feed them to NAE for the ReID task. Since online instance matching (OIM) \cite{oimnet} utilizes memory to keep the features for the labeled and unknown identities using a lookup table (LUT) and a circular queue (CQ). The LUT is defined as  $V \in \mathrm{R}^{D \times L}$  where D and L are the feature dimensions and IDs respectively, and CQ is represented as $U \in  \mathrm{R}^{D \times Q}$ where Q is the queue size. It is impractical to directly utilize the OIM for the UDA PS task. Therefore, we extended the OIM for the UDA ReID and introduced a hybrid memory projection module to keep LUT for the known source IDs and pseudo-labeled target IDs. However, we keep a single CQ for both source and target unknown identities.
Using the hybrid memory for the extended OIM,  we calculate the similarity projection $p_k$ for the input feature sample w.r.t. the LUT IDs as follows: 
\begin{equation}
    p_k = V^T_if_k, 
\label{eq_projection}
\end{equation}
where $k$ indicates the source domain, target domain, or mixed domain, and $f_k$ denotes the feature embeddings from the $k$th domain.
To quantify the discrepancy in the domain distribution between the mixed domain memory projections and the other two extreme domains, we  make employ cross-entropy loss as in Eq. \ref{b_eq_cross_entropy}. This ensures that the dynamic properties of the hybrid memory projection module  are compatible with the bridging method and allows it to work in the person search domain. We employ the $L2-Norm$ loss for the feature space to evaluate the distance across the domains (in Eq. \ref{c_eq_l2_norm}) to maintain the shortest path for the mixed feature w.r.t. the source domain and target domain.  The proposed two bridge losses are as follows:



\vspace{-0.1cm}
\begin{equation}
    \mathcal{L}^{\varphi }_{bridge}=-\frac{1}{n}\sum^{n}_{i=1}\sum_{k\in [s,t]} c^{k}_{i} \cdot [y^i_klog(p^i_{mix}))], 
\label{b_eq_cross_entropy}
\end{equation}
\vspace{-0.1cm}
\begin{equation}
    \mathcal{L}^{f}_{bridge}=-\frac{1}{n}\sum^{n}_{i=1}\sum_{k\in [s,t]} c^k_i \cdot ||f^k_i-f^{mix}_{i}||_2,
\label{c_eq_l2_norm}
\end{equation}
where $k$ represents the domain (i.e., source or target) and $i$ indicates the index in the minibatch. The $y^i_k$ shows the source label or target pseudo label, the $f_i^{k}$ denotes the $k$th domain's  representation, $p_i^{mix}$ reflects the mixed domain similarity projection, and $f_i^{mix}$ means the mixed domain features (using the $f_i^{s}$ and $f_i^{t}$), from the proposed DDAM module, respectively.

Another important property is to ensure that the mixed domain is diverse enough so that the source or the target domain does not dominate each other. To maximize the diversity of the domain constraint weights, we utilize the disparity loss. Where within the mini-batch the standard deviation $\sigma(\cdot) $ is used as follow:
\begin{equation}
    \mathcal{L}_{disp}=-[\sigma(\{c^s_i\}^n_{i=1})+\sigma(\{c^t_i\}^n_{i=1})], 
\label{d_eq}
\end{equation}
where $\sigma$ denotes the computation of standard deviations in a mini-batch. The imposed disparity loss guarantees that the mixed domain representations are as much diverse as possible to maintain the shortest geodesic path property, which can better bridge the domain gap between the source and target domains.

\subsection{Decoupled Norm-aware Embedding}\label{NAE}

As previously discussed, there is a fundamental conflict between the two subtasks, namely detection, and ReID, within the Faster RCNN-based \cite{fasterrcnn} person search frameworks. These subtasks are exposed to the same backbone network, where detection focuses on capturing common features of pedestrians, while ReID aims to discriminate the uniqueness of individuals. In fully supervised person search settings, norm-aware embeddings (NAE) take the feature vector, pass it through a shared projection layer, and decouple it into two components: norm and angle in the polar coordinate system.
However, the introduction of domain adaptation adds an additional layer of complexity to the process. Therefore, we intentionally decouple the NAE for both the detection and ReID tasks. This decoupling not only mitigates the cross-task conflict but also facilitates a more efficient handling of the ReID task in the context of UDA person search problems.


\begin{figure*}[t!]
  \centering
    \includegraphics[width=0.7\textwidth]{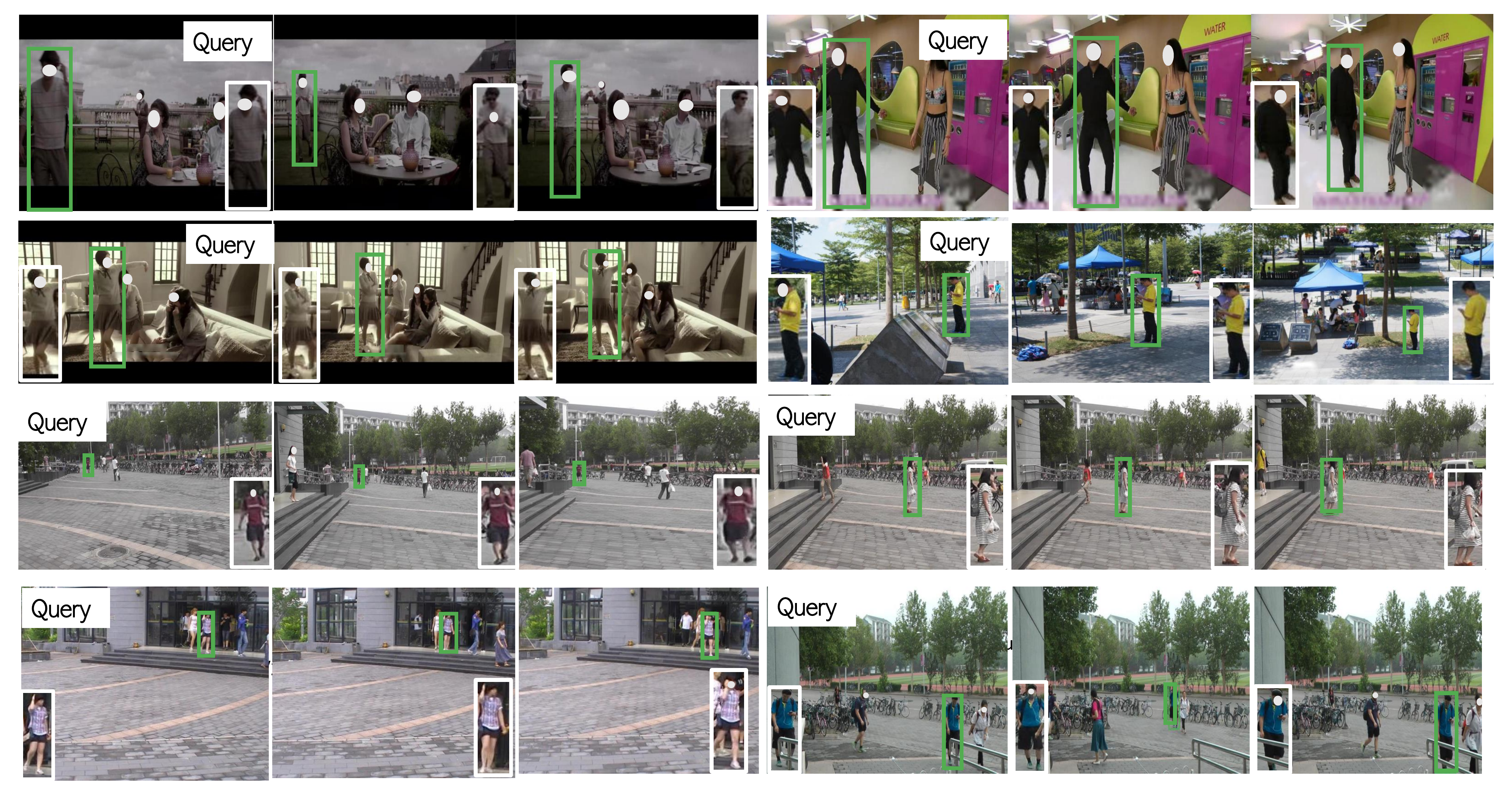}
  \caption{Qualitative analysis  on CUHK-SYSU \cite{oimnet} (top 2 rows ) and PRW \cite{PRW} (bottom  2 rows) datasets. We illustrate the top two matching results for different query persons. Our method can effectively bridge the gap using adaptive domain mixing which correctly detects and identifies.}
  \label{fig:prw_cuhk_qualitative}
\end{figure*} 

\begin{figure}[t!]
  \centering
    \includegraphics[width=\linewidth]{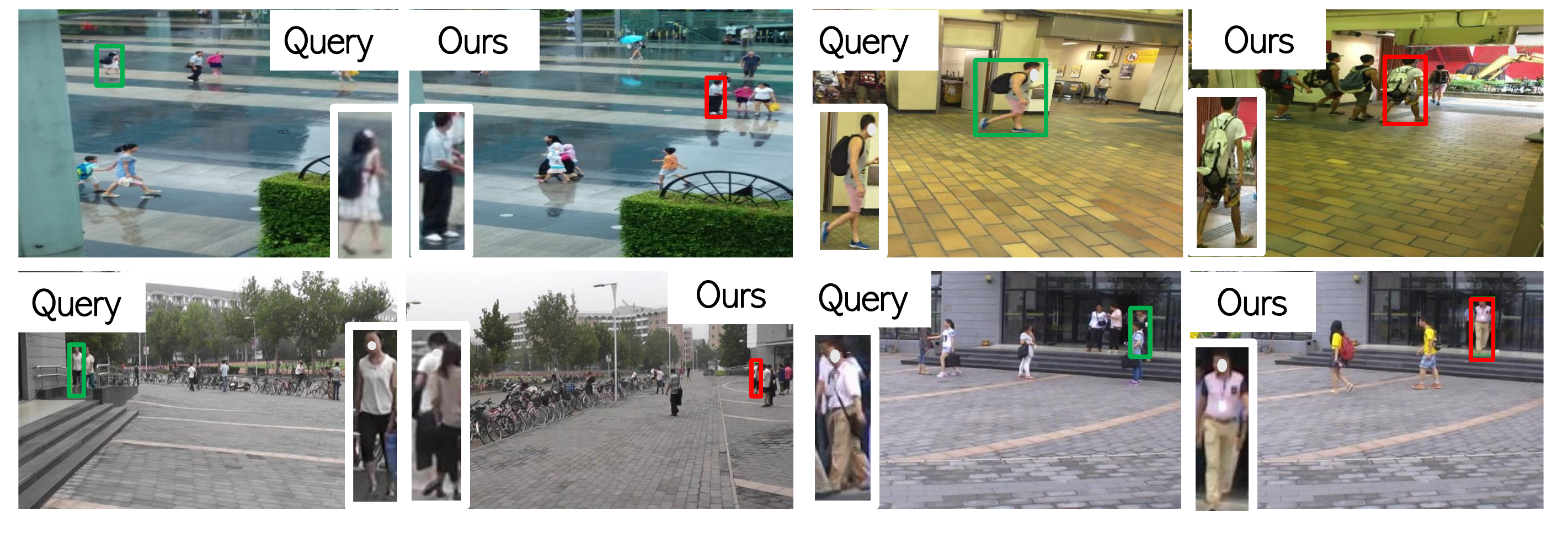}
  \caption{Failure cases on CUHK-SYSU \cite{oimnet} (first row) and  PRW \cite{PRW} (second row) datasets. We demonstrate that our approach incorrectly identifies the query person due to heavy domain conflicts between the domains.}
  \label{fig:failure_prw_qualitative}
\end{figure}

\section{Experiments}

\subsection{Implementation Details}
We implemented our method in the PyTorch framework and all experiments are performed over  NVIDIA RTX A6000 GPU. Our backbone is ResNet50 \cite{he2016deep} pre-trained over ImageNet-1K \cite{deng2009imagenet}. We resize the input to $1500 \times 900$, adopt a random horizontal flip  as augmentation, and trained our model using the Stochastic Gradient Descent (SGD) method. We train the model for 20 epochs when the target dataset is set to PRW witg batchsize of 6  and train for 10 epochs when the target dataset is set to CUHK-SYSU dataset with batch size of 4. The weight decay and momentum are set to $5 \times 10^{-4}$ and 0.9, respectively. Following \cite{daps}, we set  the learning rate 0.0024, which is reduced at epoch 16 by a factor of 0.1, and warms up at the first epoch. 
The annotations for the source domains are available during the training, whereas neither the bounding boxes of the pedestrians nor their identity information is accessible for the target domain during the training and test time. 
{Following DAPS \cite{daps}, we adopt an asynchronized training strategy and employ pseudo-bounding boxes after the $\alpha$  epochs ($\alpha$=12 for target PRW and $\alpha$=1 for CUHK-SYSU) on the target branch to supervise the box regression and classification heads. This releases the complexity of the unlabeled target domain images for both detection and ReID training. }
We utilized DDAM,to generate domain invariant representations, at the training and relinquish it during the inference time.

\subsection{Datasets and Metrics}

\noindent\textbf{Dataset}
We evaluate our method over the following two datasets, CUHK-SYSU \cite{oimnet} and PRW \cite{PRW}. 
\textbf{CUHK-SYSU}\cite{oimnet} is a large dataset for a person search with 8,432 ID individuals across 18,184 images accounting for 96,143 bounding boxes. In training, only 5,532 IDs are accessible through 11,206 images while the remaining 2900 IDs and corresponding 6,978 images are used for evaluation. The CUHK-SYSU contains two distinct data sources; 1) street view images that contain a series of variations focusing on viewpoints, lighting, resolutions, and occlusions. 2) Movies and drama serial videos that contain a variety of unique indoor and outdoor challenges. This allows the dataset to add more diversity to the scenes. For evaluation, the images are split into 2900 query persons and the 6978 images are utilized as the gallery set.
\textbf{PRW}\cite{PRW} is another dataset consisting of 932 IDs having 11,816 images with 43,110 bounding boxes. The dataset is sampled from videos that were captured from six CCTV university cameras. For training, only 482 IDs are available in 5702 images while the test set has 2057 query persons with a gallery size of 6112 images.    

\noindent\textbf{Evaluation Protocols:} For the domain adaptation setting, we evaluate our method on the test set of the target domain. In order to quantify the  localization/detection task, we use standard object detection protocols such as recall score and average precision.   We adopt widely used metrics cumulative matching characteristics (CMC) curves and mean average precision (mAP) to measure the performance of the ReID task. Since ReID reflects the identity of the query person, it is the most challenging metric for the PS task.

\begin{table*}[t]
\centering
\caption{
Ablation study on the PRW and CUHK-SYSU datasets. Here, we show the merits of our contributions introduced to the baseline (DAPS \cite{daps}). The 
$\mathcal{L}^f_{bridge}$ \& $\mathcal{L}^{\phi}_{bridge}$ represent the bridge losses, $\mathcal{L}_{disp}$ denotes the disparity loss, and  DC-NAE indicates the decoupled NAE. We note that the 
integration of our noval bridge losses (row 5) and disparity loss (row 6) leads to consistent gain in terms of mAP for both datasets. Similarly, introduced losses (row 7) and decoupled NAE (row 8) obtained better results compared to the baseline. Our final approach (row 9) achieves significant performance gain compared to the baseline and its results are in bold.}
\scalebox{0.8}{
\begin{tabular}{|l|c|c|c|c|c|c|c|c|c|c|c|c|}
\hline
  \multicolumn{5}{|c|}{Experiments} & \multicolumn{4}{|c|}{Target:PRW}   & \multicolumn{4}{c|}{Target:CUHK-SYSU}\\ \hline
Exp. No. & $\mathcal{L}^f_{bridge}$ & $\mathcal{L}^{\phi}_{bridge}$  & $\mathcal{L}_{disp}$ & DC-NAE  & mAP & Top-1 & Recall & AP & mAP & Top-1 & Recall & AP \\ \hline
 1 (baseline) & \xmark  & \xmark & \xmark & \xmark  & 34.7 & 80.6 & 97.2 & 90.9 & 77.6 & 79.6 & 77.7 & 69.9 \\ \hline 
 2 (baseline reproduced) & \xmark  & \xmark & \xmark & \xmark  & 34.4 & 78.4 & 92.1 & 87.5 & 77.1 & 78.2 & 72.8 & 67.9 \\ \hline 
 3 & \checkmark  & \xmark & \xmark & \xmark  & 34.9 & 78.9 & 92.5 & 87.6 & 77.9 & 79.2 & 73.4 & 68.1 \\ \hline 
4 & \xmark  & \checkmark & \xmark & \xmark  & 34.7 & 78.6 & 92.4 & 87.8 & 77.4 & 79.3 & 73.9 & 68.5 \\ \hline
5 & \checkmark & \checkmark & \xmark & \xmark  & 35.1 & 79.4 & 92.9 & 88.1 & 78.1 & 79.7 & 74.1 & 68.3 \\ \hline
6 & \xmark & \xmark & \checkmark & \xmark  & 35.7 & 79.0 & 92.6 & 88.0 & 78.3 & 79.8 & 74.9 & 68.1 \\ \hline
7 &  \checkmark  &  \checkmark &  \checkmark & \xmark & 35.9 & 79.5 & 92.5 & 88.4 & 78.5 & 80.7 & 75.4 & 68.7 \\ 
\hline
8 & \xmark  &  \xmark &  \xmark & \checkmark & 35.5 & 79.4 & 93.1 & 88.2 & 78.6 & 80.3 & 74.7 & 68.2 \\ 
\hline 
9   (Ours) & \checkmark  &  \checkmark &  \checkmark & \checkmark & \textbf{36.7} &\textbf{ 81.2} & \textbf{93.3} & \textbf{88.6} & \textbf{79.5} & \textbf{81.3} & \textbf{76.5} & \textbf{68.8} \\ 
\hline
\end{tabular}}
\label{tbl:ablation_study}
\end{table*}

\begin{table}[]
\centering
\caption{
A study on how efficiently the proposed methods
can adapt to a reduced-size target dataset.}
\scalebox{0.6}{ 
\begin{tabular}{|c|c|c|c|c|c|c|c|c|}
\hline
                           & \multicolumn{4}{l|}{Target:PRW} & \multicolumn{4}{l|}{Target:CUHK-SYSU} \\ \hline
Method / Sample Percentage & mAP   & Top-1  & Recall & AP   & mAP    & Top-1   & Recall   & AP     \\ \hline
Baseline / 100\%           & 34.4  & 78.4   & 92.1   & 87.5 & 77.1   & 78.2    & 72.8     & 67.9   \\ \hline
Ours / 50\%                & 32.7  & 77.6   & 91.4   & 87.0 & 76.5   & 77.1    & 72.5     & 65.9   \\ \hline
Ours / 75\%                & 34.8  & 78.9   & 92.4   & 87.3 & 77.2   & 79.0    & 74.5     & 67.8   \\ \hline
Ours / 100\%               & 36.7  & 81.2   & 93.3   & 88.6 & 79.5   & 81.3    & 76.5     & 68.8  \\ \hline
\end{tabular}}
\label{UDA_Table2}
\end{table}

\subsection{Comparison with State-of-the-art Methods}
We compared our method with fully supervised, weakly
supervised, and unsupervised domain adaption methods.
First, we present a comparison of our UDA method with the
fully supervised methods classified as two-stage and one-
stage methods in Table \ref{tab:tab_sota_comparison}. Surprisingly, our method outperforms several two-stage and one-stage fully supervised methods including DPM \cite{girshick2015deformable}, MGTS \cite{ps_mask}, OIM \cite{oimnet}, NPSM \cite{Liu_NPSM_ICCV_2017}, IAN \cite{Xiao_IAN_PR_2019}, and CTXGraph \cite{Yan_CTXG_CVPR_2019}. Second, we also compare our method with weakly supervised and UDA methods in Table \ref{tbl:comparison_with_semI_DA}. Compared to the top-performing weakly supervised method, our approach obtains an absolute gain of 13.2\% over the PRW dataset. Compared to DAPS,
our method demonstrates outstanding performance depicting the merits of our method and archives 2.0\% and 1.9\% gain in terms of mAP over both PRW and CUHK-SYSU
datasets, respectively. 

We present the qualitative comparison of our method
with DAPS in Fig. \ref{fig:daps_prw_qualitative} which depicts that our method is able
to correctly identify the query person in complex scenes.
More examples from CUHK-SYSU and PRW datasets are
shown in Fig. \ref{fig:failure_prw_qualitative}. This shows that our DDAM module facilitates correctly localizing and identifying the query person
in challenging scenarios. In Fig. \ref{fig:failure_prw_qualitative}, we also present failure
cases where there exist heavy domain differences.

\subsection{Ablation Study}
We conduct an ablation study to validate the merits of
our method in Table \ref{tbl:ablation_study}. As mentioned earlier, we adopt
DAPS [28] as our baseline. For a fair comparison, we re-
produce the baseline numbers and report in Table \ref{tbl:ablation_study} (row
2). We integrated DDAM into the baseline and trained the
model using introduced bridge losses (rows 3, 4, and 5)
and disparity loss (low 6). We notice that combined bridge and disparity losses have more gain compared to individual
bridge losses in terms of mAP for both datasets. When integrating our proposed DDAM (trained with three introduced
losses) into the baseline (row 7), the mAP score is significant improved to 35.9\% and 78.5\% in terms of mAP
over PRW and CUHK-SYSU datasets, respectively. This is
attributed to the nature of DDAM since the objective is not
to improve the quality of feature extraction but to minimize
the disparity between the two domains without a label ID
class as well as try to maintain diversity for both domains.
Similarly, the decoupling of the NAE  into the baseline (row 8) leads to improving the mAP scores over both PRW and CUHK-SYSU datasets. This is due to mitigating the issue of conflicting objectives of commoners, uniqueness, and adaption. Separating the NAE for detection and ReID eases the PS process. Finally, combining both contributions (row 9) leads to a significant improvement in performance and obtains mAP scores of 36.7\% and 79.5\% for both PRW and CUHK-SYSU datasets, respectively.

To further verify the impact of our new module, we studied how much reducing the number of training samples might affect the performance of the model. In Table \ref{UDA_Table2}, we see that the model obtains comparable results with the baseline model scores even when one-fourth of the target training set is removed (row 3).

\section{Conclusion}

We present a novel UDA person search framework that leverages a bridging mechanism to generate domain-invariant representations. Our approach introduces the DDAM module, which produces moderate mixed domain representations that effectively adapt the extremes of the two domains through an adaptive mixing mechanism, facilitating improved knowledge transfer from the source domain. To enhance the discriminability of the model on the target domain, we employ bridge and disparity losses. Additionally, we incorporate an NAE-decoupled module to mitigate the cross-task conflict, resulting in enhanced ReID quality and improved domain adaptation for the person search task. 
Our proposed contributions significantly enhance the model's domain adaptation abilities for person search. Our experimental studies validate the effectiveness of our proposed method.



\section*{Acknowledgement}
This work is partially supported by the MBZUAI-WIS Joint Program for AI Research (Project
grant number- WIS P008)

{\small
\bibliographystyle{ieee_fullname}
\bibliography{myegbib}
}

\end{document}